# Leveraging Large Language Models for Cost-Effective, Multilingual Depression Detection and Severity Assessment

Longdi Xian[a*]; Jianzhang Ni[a]; Mingzhu Wang[b]

[a]Department of Psychiatry, Faculty of Medicine, The Chinese University Of Hong Kong.

[b]Department of Linguistic and Translation, City University of Hong Kong.

First and Corresponding Author: Longdi Xian longdixian@link.cuhk.edu.hk

Co-First Author: Jianzhang Ni j.ni@link.cuhk.edu.hk

Mingzhu Wang mingzwang9-c@my.cityu.edu.hk

## Abstract

Depression is a prevalent mental health disorder that is difficult to detect early due to subjective symptom assessments. Recent advancements in large language models have offered efficient and cost-effective approaches for this object. In this study, we evaluated the performance of four LLMs in depression detection using clinical interview data. We selected the best-performing model and further tested it in the severity evaluation scenario, and knowledge-enhanced scenario. The robustness is evaluated in complex diagnostic scenarios using a dataset comprising 51074 statements from 6 different mental disorders. We found that DeepSeek-V3 is the most reliable and cost-effective model for depression detection, performing well in both zero-shot and few-shot scenarios, with zero-shot being the most efficient choice. The evaluation of the severity showed low agreement with the human evaluator, particularly for mild depression. The model maintains stably high AUCs for detecting depression in complex diagnostic scenarios. These findings highlight DeepSeek-V3's strong potential for text-based depression detection in real-world clinical applications. However, they also underscore the need for further refinement in severity assessment and the mitigation of potential biases to enhance clinical reliability.

## Introduction

Depression is one of the most pervasive mental health disorders, significantly impacting the quality of life and serving as a major cause of suicidal behavior. Despite advancements in clinical diagnostics, early and accurate detection of depression remains challenging due to the subjective nature of symptom assessment and the variability of its manifestations across different populations. In recent years, artificial intelligence (AI) and natural language processing (NLP) have emerged as promising tools to enhance traditional screening methods. In particular, Large Language Models (LLMs) built on transformer architecture have shown potential by enabling the automated analysis of linguistic data (Omar & Levkovich, 2025). Among these, BERT (Bidirectional Encoder Representations from Transformers) and its variants, such as RoBERTa and MentalBERT, have been extensively tested for depression detection. These models can identify and categorize signs of depression and have shown high capability (Acheampong et al., 2021; Bucur, 2024; Chen et al., 2024; Clusmann et al., 2023; Dai et al., 2021; Farruque et al., 2022; Sezgin et al., 2023; Toto et al., 2021; Wan et al., 2022).

One of the most notable developments in transformer-based models is the Generative Pretrained Transformer (GPT) series developed by OpenAI. Models like GPT-4o and GPT-4o mini have attracted attention for their advanced language understanding and generation capabilities, often surpassing BERT-based models in contextual comprehension, coherence, and the ability to generate human-like text (Boitel et al., 2024; Floridi & Chiriatti, 2020; OpenAI et al., 2024). These strengths make GPT models particularly promising for applications in depression detection, where nuanced interpretation of language plays a critical role (Ganesan et al., 2024; Lian et al., 2024; Lorenzoni et al., 2024; Lu et al., 2024; Tak & Gratch, 2024).

However, despite their strengths, GPT-based models have limitations; their closed-source nature and reliance on cloud-based APIs raise significant data privacy and ethical concerns in clinical settings. In contrast, newer transformer models like DeepSeek offer comparable or even superior performance while remaining open source, thereby mitigating these issues (DeepSeek-AI, Liu, et al., 2025). This opens new opportunities for effective depression symptom detection with very low costs. Recent studies have been testing the effectiveness of using DeepSeek models in depression detection(Teng et al., 2025) and simulated CBT (Tahir, 2024). These preliminary studies highlight the promise of DeepSeek models in revolutionizing depression screening by offering cost-effective, privacy-preserving diagnostic alternatives that could be seamlessly integrated into clinical practice.

Despite significant advancements in transformer-based models for depression detection, there remains a critical research gap in systematically evaluating open-source alternatives like DeepSeek that promise enhanced data privacy and cost-effectiveness in clinical settings. This study systematically evaluates the effectiveness of DeepSeek models, DeepSeek-R1 and

DeepSeek-V3, in detecting depression and assessing symptom severity. Specifically, the objectives are to: (i) Compare four modelsthe model performance with and without the prior knowledge using zero-shot and few-shot prompt schemes (ii) benchmark their performance against other large language models; (iii) examine trade-offs between predictive accuracy and computational efficiency; and (iv) explore the challenges associated with depression severity classification, including the impact of other mental disorders on diagnostic depression accuracy. Findings from this study will contribute critical insights for future research and clinical implementation of AI-assisted diagnostic tools, promoting both methodological rigor and practical applicability in mental health care.

# Methodology

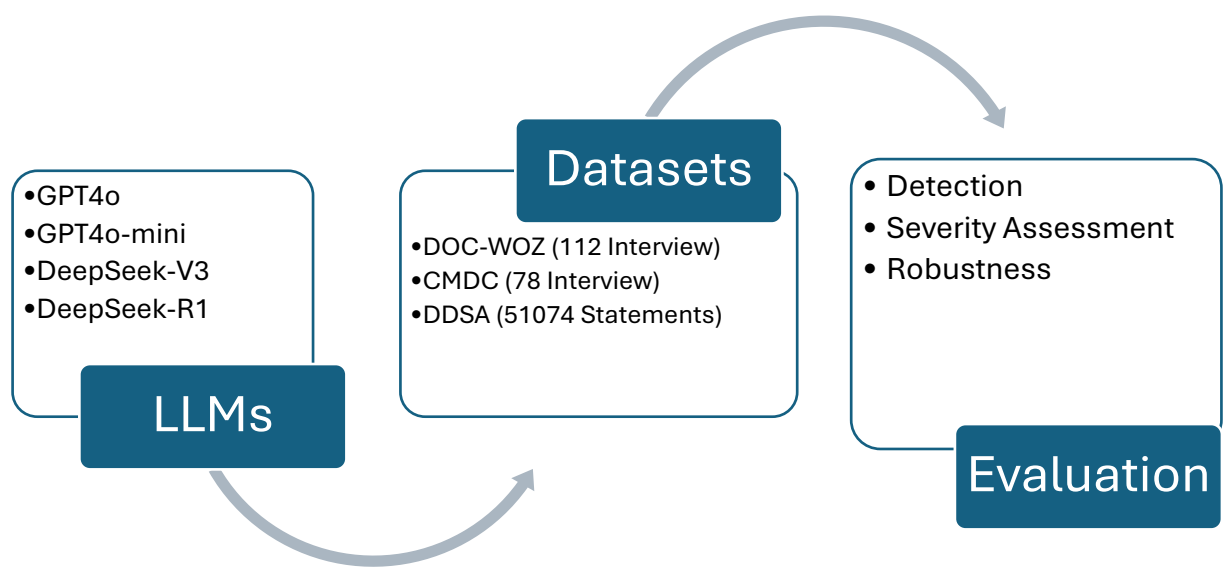

*Figure 1 Framework*

## Study Materials.

The DOC-WOZ (Depression Corpus Wizard-of-Oz), hereafter also referred to as W, is a multimodal dataset designed for depression detection, comprising data from 112 participants in English (Omar & Levkovich, 2025). It primarily focuses on speech and conversational interactions. Participants were instructed to engage in English-language conversations with a simulated AI system, using the Wizard-of-Oz paradigm—where a human operator covertly controls the AI system to simulate natural interaction. The dataset includes audio recordings, transcripts, and clinical data, such as Patient Health Questionnaire (PHQ-8) scores, which are used to assess the severity of depression.

The CMDC (Chinese Multimodal Depression Corpus), hereafter also referred to as C, contains semi-structural interviews designed to support the screening and assessment of major depressive disorder (Zou et al., 2023). It contains 26 MDD patients (8 males, 18 females) with a mean age of 24.1 (SD = 5.04, age range 19-30 years) and 52 HC individuals (17 males, 35 females) with a mean age of 30.5 (SD = 12.06, age range 20-60 years). The transcribed interview text is used for analysis.

The DDSA (Depression Detection using the Sentiment Analysis) dataset is a collection of mental health statuses in English, gathered from 51074 real statements shared by individuals from various platforms, including social media, Reddit, Twitter, and others. The statements are categorized under the following 7 categories, including Normal (16351), Depression (15,404), Suicidal (10,653), Anxiety (3,888), Bipolar (2,877), Stress (2,669), and Personality Disorder (1,210). This large and diverse dataset enables comprehensive evaluations in the real-world scenario. The details of the included datasets are shown in Table 1.

*Table 1 Dataset information*

| | DOC-WOZ (W) | | | CMDC (C) | | DDSA | |
|---|---|---|---|---|---|---|---|
| **Normal** | 94 | | | Normal | 52 | Normal | 16351 |
| **Depression** | 125 | mild | 82 | Depression | 126 | Depression | 15404 |
| | | | | | | Suicidal | 10653 |
| | | moderate | 25 | | | Anxiety | 3888 |
| | | | | | | Bipolar | 2877 |
| | | severe | 18 | | | Stress | 2669 |
| | | | | | | Personality Disorder | 1210 |

## Large Language Models and Prompt Engineering

Four LLMs are evaluated in this study, namely, GPT-4o, GPT-4o-mini, DeepSeek-V3, and DeepSeek-R1. The R1 model is optimized for high precision and reliability, effectively identifying subtle emotional and cognitive patterns in text, whereas the V3 model emphasizes versatility and ease of use, supporting real-time processing and personalized insights. These capabilities have been particularly beneficial in mental health research, such as detecting depression via linguistic analysis (Zou et al., 2023). An example of the parameters and prompts used is shown in Table 2, with adjustments according to the experiment setup described below.

*Table 2 Prompt Engineering*

| Component | Description | Formula/Value |
|---|---|---|
| Temperature | - | 0.2 |
| Input Labels | A subset of labels provided for each classification task. | label_subset = ["Normal", "Depression", "Suicidal", ...] |
| JSON Structure | Labels mapped to probabilities that sum to 1.0. | $\{"label_1": s_1, "label_2": s_2, ..., "label_i": s_i\}$ where $\Sigma(s_1^i) = 1.0$ |
| Prompt Template | Instruction for the model to classify the statements. | You are a mental health classifier. For each statement, output probabilities for these labels in JSON format: {...} |
| Label Format | Labels converted to JSON format with initial probabilities set to 0.0. | ",\n".join([f'"{label}": 0.0' for label in label_subset]) |
| Example Prompt Output | The prompt dynamically generated for a label subset (e.g., ["Normal", "Depression"]). | You are a mental health classifier, output probabilities for these labels in JSON format: {"Normal": 0.0, "Depression": 0.0 } Only respond with valid JSON. The probabilities for the provided |

## Experiment Setup

We first tested and compared different LLMs' performance using DOC-WOZ and CMDC datasets and finally evaluated the performance of the selected model with DDSA. In detail, the dataset can be denoted as

$$D_i^n = \{(s_1, \dots, s_n)\}, \forall i \epsilon [1,3], n \epsilon [0,6], \quad (1)$$

Where n was the which labels (0:no depression;1…), $s_n$ was the $n$-th label in the dataset (details shown in Experiments), and $D_i^n$ was the include n labels $i$-th dataset (details shown in dataset session).

Based on these datasets, we used different LLMs in same $i$-th dataset in same $n$-th statement using Key metrics such as the Area Under the Curve (AUC), precision (PPV), negative predictive value (NPV), sensitivity, and specificity were used to evaluate detection depression accuracy. And using pricing and times as basic evaluation standard.

$$E_\partial^\alpha = \begin{cases} \partial: AUC = \Sigma \left[ \frac{(F_{i+1} - F_i) \times (T_{i+1} + T_i)}{2} \right] \\ \dots \dots \\ \alpha: sp = \frac{TN}{TN + FP} \end{cases}, \quad (2)$$

Were $F_i$ was $i$-th the vertical axis is TPR (False Positive Rate). The $T_i$ $i$-th horizontal axis is TPR (True Positive Rate) for the ROC curve chart. The True Negatives ($TN$): The number of samples correctly predicted as negative by the model. False Positives ($FP$): The number of samples in the negative category that were incorrectly predicted by the model to be positive (negative).

$$B_\beta = \left( P_\$^s \cap E_\partial^\alpha \left( M_1^j \{D_i^n\} \right) \right), \forall n \epsilon [0,1], \quad (3)$$

Where $M_1^j$ was the from 1 to $j$-th LLM in include 0(no depression) and 1(depression) labels $i$-th dataset and $E_1^\alpha$ was used from $\partial$ to $\alpha$ evaluate metrics. And the $\partial, \alpha$ stand for different evaluation metrics. $P_\$^s$ was the pricing ($) and time ($s$) as basic assessment methods. $\beta$ was the $\beta$-th LLM have the best performance.

Secondly, the study was designed to systematically evaluate the performance of the best model ($B_\beta$) in detecting depression across different learning paradigms, with a focus on zero-shot and few-shot settings.

$$f_\vartheta = \varepsilon_{\gamma \leftarrow D_i^{n=1} | \gamma} \left( \left( E_\partial^\alpha (B_\beta) \right) \right), \quad (4)$$

Where $\gamma$ were the zero-shot and $\gamma \leftarrow D_i^{n=1}$ the few shots referred to knowledge from another dataset. And then, $\varepsilon$ there were zero shots or few shots in process the that evaluated the LLM in both performances. $f_\vartheta$ was the best method in zero or few shots in the detection of depression.

This approach tests the models' adaptability when have another dataset as knowledge or do not have other examples as reference. The performance differences between zero-shot and few-shot configurations will be analyzed to determine whether limited supervision significantly enhances detection reliability based on accuracy metrics ($E_\partial^\alpha$).

Thirdly, building on these findings, the next phase examines the models' ability to discern varying depression severity levels (1: mild, 2: moderate, and 3: severe) using the performance of the best model ($B_\beta$).

$$\beth_\omega = E_\partial^\alpha\left(C_1^{\mu \leftarrow D_y^n}(B_\beta)\right), \forall \mu \epsilon [2,3], \forall n \epsilon [0,3] \qquad (5)$$

Where $C_1^\mu$ was from $D_y^n$ dataset selects $\mu$ labels data in depression different levels as experiment samples. And the $\beth_\omega$ was $\omega$ labels was the best performance. This stratified analysis will reveal whether detection sensitivity scales with clinical urgency, are a critical factor for triage or intervention planning.

Finally, the study examines the potential influence of diagnostic heterogeneity by testing on DDSA dataset which contains six distinct mental disorders. This evaluation aims to determine whether the presence of non-depression diagnoses interferes with the model's ability to accurately classify depression in a multi-label, non-binary setting.

$$\tau_\vartheta = E_\partial^\alpha\left(C_1^{s \leftarrow D_m^n}(B_\beta)\right), \forall s \epsilon [2,5] \; \forall n \epsilon [0,6], \qquad (6)$$

Where $C_1^s$ was from $D_m^n$ dataset select $s$ labels data in different mental disorders to add depression and normal as experiment samples. And the $\tau_\vartheta$ way $\vartheta$ labels was the best performance. By comparing depression detection accuracy in isolated versus other mental disorders' cases, we aim to uncover biases or robustness in the models' diagnostic logic.

## Result

### Individual Models' Performance

This table provides a comparative performance analysis of four large language models (LLMs): ChatGPT-4o, ChatGPT-4o mini, DeepSeek R1, and DeepSeek V3, evaluated on two different main used language datasets—DOC-WOZ (English) and CMDC (Chinese). The metrics used include AUC (Area Under Curve), Accuracy, PPV (Positive Predictive Value), NPV (Negative Predictive Value), Sensitivity, Specificity, processing time (in minutes), and pricing (in USD).

*Table 3 Individual LLM performance in ChatGPT and Deepseek in different languages' datsets*

| | GPT4o | GPT4o-mini | Deepseek-R1 | Deepseek-V3 |
|---|---|---|---|---|

| Dataset | W | C | W | C | W | C | W | C |
|---|---|---|---|---|---|---|---|---|
| AUC | 0.83*** | 0.99*** | 0.83*** | 0.98*** | 0.82*** | 0.98*** | 0.83*** | 0.98*** |
| Accuracy | 0.74 | 0.83 | 0.60 | 0.92 | 0.74 | 0.85 | 0.71 | 0.85 |
| PPV | 0.88 | 0.76 | 0.93 | 0.83 | 0.84 | 0.68 | 0.86 | 0.68 |
| NPV | 0.61 | 1.00 | 0.52 | 0.97 | 0.65 | 1.00 | 0.62 | 1.00 |
| Sensitivity | 0.58 | 1.00 | 0.34 | 0.96 | 0.66 | 1.00 | 0.59 | 1.00 |
| Specificity | 0.89 | 0.85 | 0.97 | 0.90 | 0.83 | 0.77 | 0.87 | 0.77 |
| Times (seconds) Per case | 1.33 | 1.03 | 1.01 | 1.23 | 25.59 | 20.21 | 6.25 | 6.16 |
| Pricing (Total) (US $) | 1.72 | 0.54 | 0.53 | 0.16 | 0.10 | 0.04 | 0.08 | 0.025 |

Note: *** stands for p<.001. p-value calculated using the DeLong ROC test.

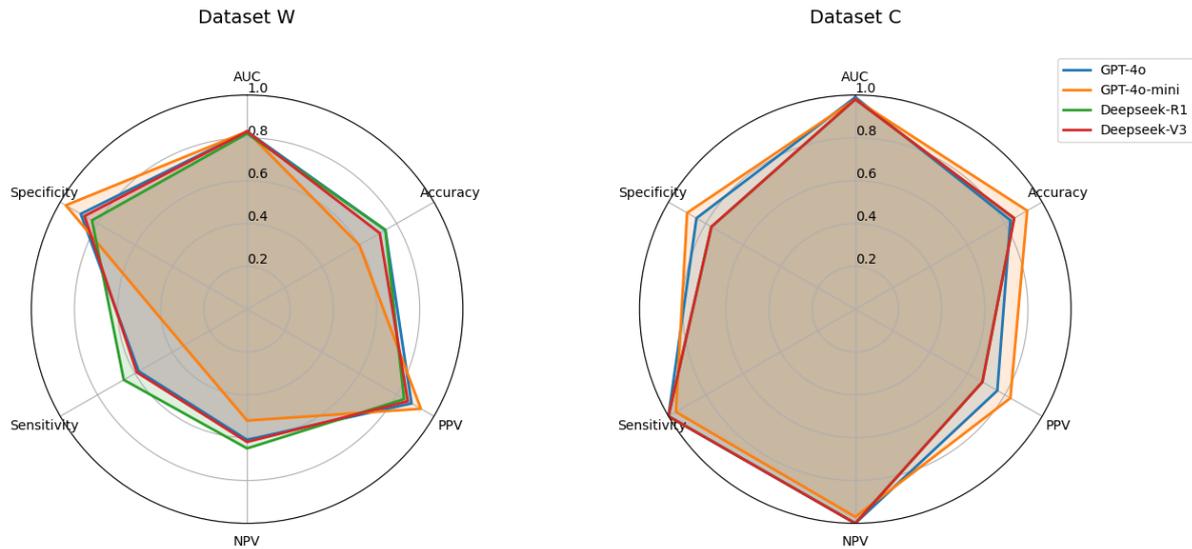

Figure 2 Radar Chart For performance in depression detection within ChatGPT and Deepseek

For Dataset W (DOC-WOZ, English), model performance is largely consistent across all evaluated LLMs. The AUC values are nearly identical at 0.82–0.83, with p < .001 indicating high statistical significance in classification performance across the models. Accuracy ranges from 0.60 (GPT-4o-mini) to 0.74 (GPT-4o and Deepseek-R1), and all models demonstrate solid specificity (⩾ 0.83). GPT-4o-mini shows the lowest sensitivity (0.34), while Deepseek-R1 achieves the highest (0.66), suggesting a better ability to detect positive cases. From a computational efficiency standpoint, the inference time per case ranges from 1.01 seconds (GPT-4o-mini) to 25.59 seconds (Deepseek-R1). Deepseek-V3 offers a strong middle ground, taking 6.25 seconds per case, significantly faster than Deepseek-R1 but slower than ChatGPT models. In terms of total cost for processing the dataset, DeepSeek models are clearly more economical. DeepSeek -R1 costs $0.10 and Deepseek-V3 just $0.08, compared to $1.72 for GPT-4o and $0.53 for GPT-4o-mini. This substantial cost reduction makes DeepSeek models attractive for large-scale English-language applications.

For Dataset C (CMDC, Chinese), all models perform at a very high level, with AUCs between 0.98 and 0.99 (p < .001). Sensitivity is extremely high across the board, with three of the four models achieving perfect sensitivity (1.00). Accuracy is highest for GPT-4o-mini (0.92), with other models also performing well between 0.83 and 0.85. Negative Predictive Value (NPV) is perfect (1.00) for all models except GPT-4o-mini, which still performs strongly at 0.97. The inference time per case remains low for all models in Dataset C, ranging from 1.03 seconds

(GPT-4o) to 20.21 seconds (Deepseek-R1). Deepseek-V3 achieves near ChatGPT speeds at 6.16 seconds per case. Crucially, total cost for processing the full dataset in Chinese is dramatically lower for DeepSeek: only $0.04 for Deepseek-R1 and $0.025 for Deepseek-V3, compared to $0.54 for GPT-4o and $0.16 for GPT-4o-mini.

Across both datasets, Deepseek-V3 offers the best overall trade-off between performance, processing time, and cost. It matches the AUC, sensitivity, and specificity of larger models like GPT-4o, while cutting costs by over 90% and keeping inference time within a reasonable range. These advantages make Deepseek-V3 particularly suited for practical deployment in both English (DOC-WOZ) and Chinese (CMDC) applications, especially when budget and scalability are key concerns.

Knowledge Enhanced Depression Detection using DeepSeek-V3

Table 4 Deepseek-V3 performance in detection depression within Zero shot and few Shot

| | DeepSeek-V3 | | | |
| --- | --- | --- | --- | --- |
| | Zero Shot | | Few Shot | |
| Dataset | W | C | W_C | C_W |
| AUC | 0.83*** | 0.98*** | 0.83*** | 0.99*** |
| Accuracy | 0.71 | 0.85 | 0.74 | 0.74 |
| PPV | 0.86 | 0.68 | 0.83 | 0.57 |
| NPV | 0.62 | 1.00 | 0.66 | 1.00 |
| Sensitivity | 0.59 | 1.00 | 0.69 | 1.00 |
| Specificity | 0.87 | 0.77 | 0.82 | 0.62 |
| Times (Total) (minutes) | 21 | 13 | 63.4 | 41 |
| Pricing (Total) (US $) | 0.08 | 0.025 | 1.68 | 0.8 |

Note: C_W refers to the CMDC dataset using DOC-WOZ as its knowledge base, while W_C refers to the DOC-WOZ dataset using CMDC as its knowledge base. *** stand for p<.001.

The few-shot models (W_C and C_W) show slight improvements over the zero-shot models (W and C), particularly in metrics like AUC, sensitivity, and cost, but these differences are not statistically significant (p = 0.25). For example, W (zero-shot) has an AUC of 0.83, while W_C (few-shot) achieves a higher AUC of 0.99. Similarly, C (zero-shot) also shows an AUC of 0.83, with C_W (few-shot) reaching 0.99. These improvements in AUC suggest better overall model accuracy with the few-shot approach, but the p-value of 0.25 indicates that the difference is not large enough to be considered statistically significant.

In terms of other metrics, accuracy is relatively similar between the zero-shot and few-shot models. For Dataset W (DOC-WOZ, English), the zero-shot model has an accuracy of 0.71, while W_C (Few Shot) has 0.74. For Dataset C (CMDC, Chinese), both the zero-shot and few-shot models have an accuracy of 0.74. The sensitivity for the few-shot models (W_C and C_W) is perfect (1.00) compared to the zero-shot models, with W at 0.59 and C at 0.69. However, the specificity for C_W (0.62) is slightly worse than C (0.87), and for W_C (0.59) compared to W (0.86). When considering cost and time efficiency, the few-shot models show improvements. For example, W_C takes 41 minutes to process and costs $0.80, compared to W, which takes

21 minutes and costs $0.08. Similarly, C_W is faster (41 minutes vs. 63.4 minutes) and cheaper ($0.80 vs. $1.68) than the zero-shot model C. Despite these minor improvements in processing time and cost for the few-shot models, the lack of statistical significance (p = 0.25) suggests that the zero-shot models are sufficiently effective. The differences in performance do not justify switching to the few-shot approach for most practical applications, especially considering the added complexity and cost of the few-shot models.

Depression Severity Evaluation with DeepSeek-V3

Table 5 **Performance of DeepSeek-V3 in Detecting Depression and Assessing Depression Severity**

| Type | Detection | Severity | | |
| --- | --- | --- | --- | --- |
| | | Mild | Moderate | Severe |
| AUC | 0.83*** | 0.61** | 0.69*** | 0.77*** |
| Accuracy | 0.71 | 0.63 | 0.89 | 0.92 |
| PPV | 0.86 | 0.99 | 0.00 | 0.00 |
| NPV | 0.62 | 0.63 | 0.89 | 0.92 |
| Sensitivity | 0.59 | 0.01 | 0.00 | 0.00 |
| Specificity | 0.87 | 1.00 | 1.00 | 1.00 |

*Note: \*\*\* stand for p< .001*

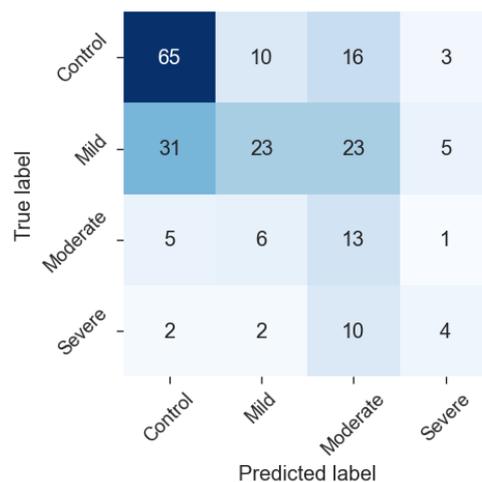

*Figure 3 Confusion Matrix with binary and multi labels for depression Detection*

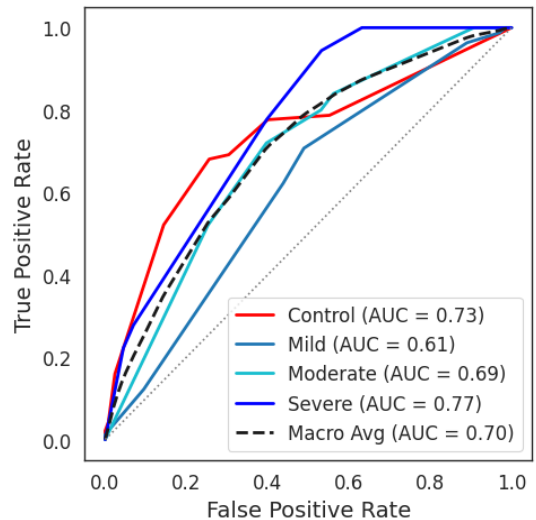

*Figure 4 2 AUC with binary and multi labels for depression Detection*

Based on Table 5, we found that the performance of DeepSeek-V3 in binary classification (i.e., distinguishing between depressed and non-depressed individuals) demonstrates strong and statistically significant results. The model achieved an AUC of 0.83 (p < .001), indicating excellent discriminative power in identifying depression in a binary setting. Additionally, the model shows reasonable accuracy (0.71), with high positive predictive value (PPV = 0.86) and specificity (0.87), suggesting that when the model predicts depression, it is usually correct, and it rarely misclassifies non-depressed individuals as depressed. While sensitivity is moderate at 0.59—meaning it detects just over half of actual depression cases—the overall balance of metrics confirms that DeepSeek-V3 performs reliably in a binary detection scenario.

In contrast, the performance of DeepSeek-V3 in multi-class classification, where the goal is to distinguish between mild, moderate, and severe depression, is considerably weaker. Although the AUCs for moderate (0.69) and severe (0.77) depression are statistically significant (p < .001), they do not translate into meaningful clinical performance. Most notably, the sensitivity for all three depression levels is either extremely low (0.01 for mild) or completely absent (0.00 for moderate and severe), indicating that the model fails to detect true cases of each severity level. While the model reports high accuracy for moderate and severe classes (0.89 and 0.92, respectively), this is likely due to class imbalance and overprediction of negative cases, as reflected in the perfect specificity (1.00) but zero PPV for these categories. In essence, the model rarely predicts moderate or severe depression, and when it does, those predictions are incorrect.

This stark contrast between binary and multi-class performance highlights a significant limitation of DeepSeek-V3: while it is well-suited for detecting depression in general, it struggles to differentiate between the severity levels. The differences in AUC between binary classification and the individual severity classes (with p < .001) further confirm that the decline in performance is statistically significant.

**Evaluation of Model Robustness with Additional Mental Health Categories**

Table 6 *Model Performance Across Varying Levels of Clinical Heterogeneity in DeepSeek-V3*

|  | I | II | III | IV | V | VI |
|---|---|---|---|---|---|---|
| **AUC** | 0.97*** | 0.96*** | 0.95*** | 0.96*** | 0.96*** | 0.97*** |
| **Accuracy** | 0.86 | 0.88 | 0.79 | 0.76 | 0.73 | 0.72 |
| **PPV** | 0.79 | 0.89 | 0.95 | 1.00 | 0.98 | 1.00 |
| **NPV** | 0.97 | 0.87 | 0.71 | 0.68 | 0.65 | 0.64 |
| **Sensitivity** | 0.98 | 0.87 | 0.60 | 0.52 | 0.47 | 0.44 |
| **Specificity** | 0.74 | 0.89 | 0.97 | 1.00 | 0.99 | 1.00 |

*Note: Columns I through VI reflect the model's performance in classifying depression under progressively more heterogeneous datasets, starting with depression and control cases (I), adding suicidal data (II), incorporating anxiety (III), adding bipolar disorder (IV), including stress (V), and introducing personality disorder (VI).*

DDSA dataset represents a substantial resource containing text from multiple mental disorders from the real world. The evaluation results are shown in Table 6, which demonstrates that DeepSeek sustains robust performance in detecting depression even as the dataset is getting complex, and more diagnostic options are added to the prompt. The consistently high AUC values (ranging from 0.95 to 0.97) indicate that the model maintains near-perfect discrimination between depressed and non-depressed individuals regardless of the increased complexity in the input data. Although there is a minor fluctuation slight dip to an AUC of 0.95 in one condition—this variation is neither statistically nor clinically significant, confirming that the addition of more data and prompt options does not compromise the model's core effectiveness.

Other performance metrics, however, reveal interpretable trends associated with the increased dataset complexity. Accuracy declines progressively from 0.86 in the initial dataset to 0.72 when more data and additional classification options are introduced. This decrease is expected, as the classification task becomes inherently more challenging with a broader range of data inputs. Conversely, precision (PPV) improves dramatically—from 0.79 in the simpler scenario to 1.00 in the expanded conditions, indicating that when DeepSeek-V3 predicts depression with the richer dataset, its positive predictions are almost always correct. Specificity also reaches 1.00 in these cases, reflecting a minimal rate of false positives.

However, these enhancements in precision come with a trade-off in sensitivity, which declines sharply from 0.98 to 0.44 as the dataset complexity increases. This suggests that while the model becomes more selective and confident in its positive predictions, it may miss a higher proportion of true depression cases when handling more complex inputs. The decrease in negative predictive value (NPV), from 0.97 to 0.64, further underscores this trend, highlighting the challenges in confidently ruling out depression when additional data and prompt options are considered.

### Discussion

The results of this study highlight DeepSeek-V3 as the most practical and efficient choice for depression detection across both English (DOC-WOZ) and Chinese (CMDC) datasets. While its AUC scores (0.83 for English, 0.98 for Chinese) are nearly identical to those of GPT-4o, DeepSeek-V3 drastically reduces cost—only $0.08 and $0.025, compared to GPT-4o's $1.72 and $0.54. Its slightly slower processing times (around 6 seconds per case) are a reasonable trade-off considering the significant savings and consistent high performance across languages. What this means for the reader is that DeepSeek V3 offers the best balance of accuracy, affordability, and multilingual robustness, making it the ideal choice for deploying depression detection models in real-world, large-scale applications.

In terms of zero-shot learning, the findings reveal that DeepSeek-V3 performs well in this approach, with AUC scores of 0.83 for both English and Chinese datasets. The slight improvements observed in few-shot learning (with AUC scores of 0.99 for both datasets) suggest that the model benefits from additional training data, but these improvements were not statistically significant. The results indicate that the zero-shot models are sufficiently effective, and the few-shot models, while marginally better, do not provide a compelling reason to shift from zero-shot learning for most practical applications. This supports the conclusion that DeepSeek-V3 can effectively detect depression even without additional training data, making it a versatile and practical solution.

DeepSeek-V3 performs well in depression detection, achieving a high AUC of 0.83 ($p < .001$) with strong accuracy and precision. However, it shows substantial differences with human evaluators when evaluating the severity. This contrast highlights a key limitation: while effective for general screening, DeepSeek-V3 currently lacks the ability to evaluate depression severity as the human evaluator, which is crucial for clinical decision-making.

The stability of the AUC confirms that DeepSeek-V3's ability to discriminate between depressed and non-depressed individuals remains strong despite the enriched dataset and expanded prompt options. Although there is a reduction in overall accuracy and sensitivity in more complex settings, the improvements in PPV and specificity indicate a favorable trade-off, ensuring that positive depression diagnosis are highly reliable. These findings support the conclusion that DeepSeek-V3 remains a dependable tool for depression screening across varied data conditions, albeit with an inherent trade-off between precision and recall in more nuanced classification scenarios.

## Limitation

While the results of this study are promising, several limitations should be addressed in future work. First, feature extraction or fine-tuning strategies should be explored to improve the model's sensitivity, particularly in the detection of mild and moderate depression, as the model's performance in these areas was suboptimal. Finally, real-world validation through clinical trials with diverse populations is necessary to assess the model's reliability outside structured datasets and ensure its applicability in hospital settings.

## Conclusion

In conclusion, DeepSeek-V3 proves to be the most reliable and cost-effective model for depression detection across both English and Chinese datasets. It delivers strong performance in both zero-shot and few-shot scenarios, achieving high AUC scores and efficient processing times, making it highly suitable for practical applications. Although GPT-4o showed slightly better AUC in some cases, its significantly higher cost makes it less practical for widespread use. DeepSeek-V3 performs well in binary detection but struggles with severity classification, especially for mild depression. Furthermore, DeepSeek-V3 demonstrated resilience in detecting depression in the complex scenario. In future work, improving sensitivity for severity detection and real-world validation through clinical trials are key areas to address.